\documentclass[conference]{IEEEtran}
\usepackage{times}

\usepackage{pdfpages}
\usepackage[numbers]{natbib}
\usepackage{multicol}
\usepackage{tikz}
\usepackage{graphicx}
\usepackage{amsmath, amsfonts, amsthm, amssymb}
\usepackage[ruled,norelsize]{algorithm2e}
\usepackage{subcaption}
\usepackage{caption}% <-- added
\usepackage{tabulary}
\usepackage[para]{threeparttable}
\usepackage{array,booktabs,longtable,tabularx}
\newcolumntype{L}{>{\raggedright\arraybackslash}X}% <-- added
\usepackage{ltablex}% <-- added
\usepackage{siunitx}% <-- added
\LinesNumbered
\usepackage{pifont}
            
\DeclareMathOperator*{\argmin}{argmin}
\DeclareMathOperator*{\argmax}{argmax}

\newcommand{\norm}[1]{\left\lVert#1\right\rVert}

\begin{document}

\title{Manipulation Trajectory Optimization with Online Grasp Synthesis and Selection}

% You will get a Paper-ID when submitting a pdf file to the conference system
% \author{Author Names Omitted for Anonymous Review. Paper-ID [66]}

\author{
Lirui Wang\textsuperscript{\rm 1},
Yu Xiang\textsuperscript{\rm 2},
Dieter Fox\textsuperscript{\rm 1,2}\\
\textsuperscript{\rm 1}University of Washington,
\textsuperscript{\rm 2}NVIDIA\\
\small{\texttt{liruiw@cs.washington.edu}, \texttt{yux@nvidia.com}, \texttt{fox@cs.washington.edu}}
% Anonymous Authors \\ Paper ID 234  
}

%\author{\authorblockN{Michael Shell}
%\authorblockA{School of Electrical and\\Computer Engineering\\
%Georgia Institute of Technology\\
%Atlanta, Georgia 30332--0250\\
%Email: mshell@ece.gatech.edu}
%\and
%\authorblockA{Twentieth Century Fox\\
%\and
%\authorblockN{James Kirk\\ and Montgomery Scott}
%\authorblockA{Starfleet Academy\\
%San Francisco, California 96678-2391\\

% avoiding spaces at the end of the author lines is not a problem with
% conference papers because we don't use \thanks or \IEEEmembership

%Homer Simpson\authorrefmark{2},
%Montgomery Scott\authorrefmark{3} and
%Eldon Tyrell\authorrefmark{4}}
%Telephone: (800) 555--1212, Fax: (888) 555--1212}
%\authorblockA{\authorrefmark{4}Tyrell Inc., 123 Replicant Street, Los Angeles, California 90210--4321}}

\maketitle

\begin{abstract}

In robot manipulation, planning the motion of a robot manipulator to grasp an object is a fundamental problem. A manipulation planner needs to generate a trajectory of the manipulator to avoid obstacles in the environment and plan an end-effector pose for grasping. While trajectory planning and grasp planning are often tackled separately, how to efficiently integrate the two planning problems remains a challenge. In this work, we present a novel method for joint motion and grasp planning. Our method integrates manipulation trajectory optimization with online grasp synthesis and selection, where we apply online learning techniques to select goal configurations for grasping, and introduce a new grasp synthesis algorithm to generate grasps online. We evaluate our planning approach and demonstrate that our method generates robust and efficient motion plans for grasping objects in cluttered scenes.

% \footnote{Our code can be found at \link{https://github.com/liruiw/OMG} and video can be viewed at \link{https://www.youtube.com/watch?v=LIcACf8YkGU}.}

\end{abstract}

\IEEEpeerreviewmaketitle

\section{Introduction}

In order to manipulate objects in a cluttered scene, a robot needs to plan its motion to reach and grasp a target object in the scene. Due to the complexity of this manipulation planning problem, it is usually decomposed into two sub-problems and tackled separately: arm motion planning and grasp planning. While arm motion planning aims at reaching the target goal and avoiding obstacles in the scene, grasp planning generates feasible grasps of the target object based on its properties such as its 3D shape and material. However, separating motion planning and grasp planning requires some post-processing steps to connect the two plans. For instance, a common strategy is to loop over each planned grasp and check if there exists a trajectory to reach the grasp. The post-processing usually involves heuristics and leads to sub-optimal solutions. How to efficiently integrate motion planning and grasp planning remains a challenge.

Solutions to this problem are previously introduced both in the traditional motion planning literature \cite{dragan2011learning,vahrenkamp2010integrated} and the recent Task And Motion Planning (TAMP) literature \cite{lagriffoul2014efficiently,toussaint2018differentiable}. According to when grasps are generated during planning, these methods can be classified into two categories. Methods in the first category generate a fixed set of grasps offline, and formulate the problem as a grasp selection problem \cite{berenson2011task,dragan2011learning,dragan2011manipulation}. Generating grasps offline can take advantages of well-designed grasp planning algorithms such as \cite{miller2004graspit}. However, a fixed set of grasps limits the number of grasp options during motion planning. Methods in the second category formulate and solve a joint optimization problem of trajectories and grasps, which generate grasps online during trajectory planning \cite{vahrenkamp2010integrated,fontanals2014integrated,haustein2017integrating}. Online grasp planning explores more grasp options. However, since the space of possible grasps is large and highly non-convex for various objects, approximations are often made to simplify the grasp planning problem. For instance, \cite{vahrenkamp2010integrated,toussaint2018differentiable} use spheres to approximate the object shape in optimizing grasps, which limits the quality of the generated grasps.

\begin{figure} 
\centering
\begin{center} 
\includegraphics[width=0.4\textwidth]{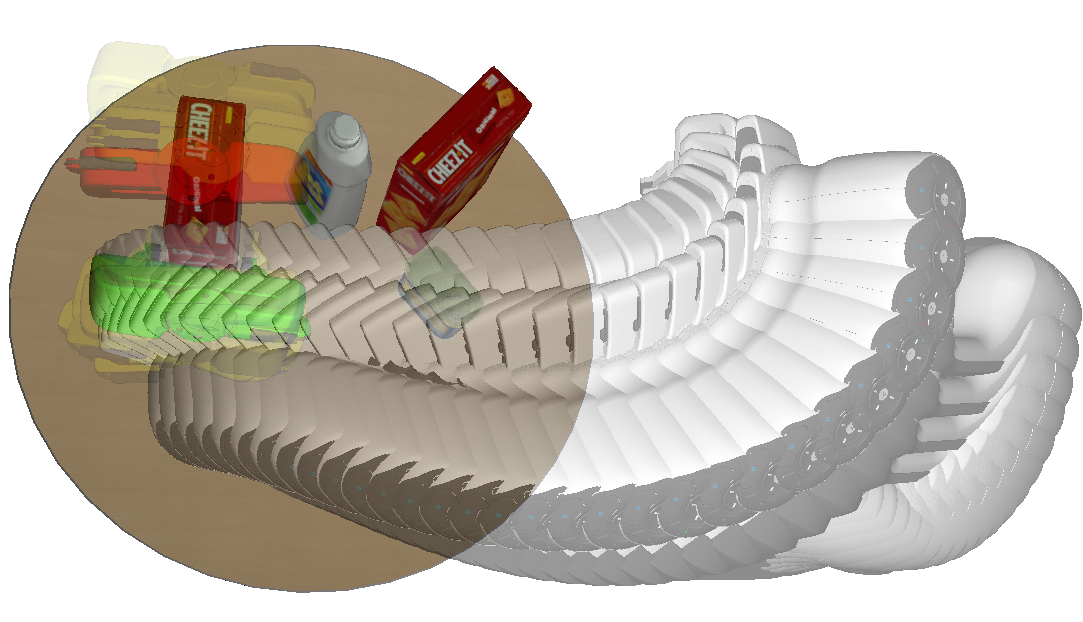}
\caption{A planning scene example. The goal set, initial grasp and selected grasp are yellow, red and green, respectively. Our algorithm synthesizes and selects grasps online and generates motion trajectories to reach the selected grasp.}
\label{fig:intro}
\vspace{-4mm}
\end{center}
\end{figure}

In this work, to overcome the limitations of existing integrated motion and grasp planning methods \cite{vahrenkamp2010integrated,fontanals2014integrated,haustein2017integrating}, we introduce a novel integrated Optimization-based Motion and Grasp Planner (OMG Planner). Our planner combines trajectory optimization with online grasp synthesis and selection. In this context, the planning phase is offline, and we do not consider execution or re-planning. ``Online'' mainly attributes to the fact that the final grasp is not fixed and can be refined during trajectory optimization. Grasp selection enables us to leverage dedicated grasp planning methods \cite{miller2004graspit,clemen2019} to generate an initial grasp set, which guarantees the quality of the grasps and handles objects with arbitrary shapes. By performing online grasp synthesis, our method augments the initial grasp set and provides more grasp candidates during motion planning. More importantly, the grasp selection problem is formulated as an online learning problem, where a probability distribution of the grasp set is estimated and updated during trajectory optimization, and the best grasp is selected accordingly.

% Instead of optimizing trajectories and grasps from scratch, we leverage existing grasp planning methods to generate an initial grasp set.  for online grasp selection our method guarantees the quality of the selected grasps. In addition, online learning techniques can be applied for grasp selection. By performing online grasp synthesis, our method augments the initial grasp set and provides more grasp candidates during motion planning.

Specifically, our planner optimizes for motion trajectories and grasps jointly. An iterative update algorithm is introduced that alternatively updates trajectories and grasps. In each iteration, given a grasp goal configuration, we utilize the CHOMP trajectory optimization method \cite{ratliff2009chomp} with a goal set constraint \cite{dragan2011manipulation} to update the trajectory. To update the grasp, we first assume the availability of a set of grasp candidates that can be obtained from any grasp planning method. Our online grasp selection algorithm then leverages the interplay between the grasp set and the current trajectory to update the probability distribution of the grasp set, from which the optimal goal in the grasp set can be selected. In order to augment the grasp set, we propose a grasp synthesis procedure named Configuration Space Iterative Surface Fitting (C-Space ISF) that  refines the grasp configuration of the selected grasp by matching the 3D shapes of the robot gripper and the object. The overall framework leverages the grasp set structure to improve motion and grasp generation. Fig.~\ref{fig:intro} illustrates a planning scene example from our method.

We conduct qualitative and quantitative evaluation to demonstrate the performance of our method for manipulation planning. We also conduct ablation studies on each component of the method. Overall, our contributions are: 
\begin{enumerate}
    \item We introduce an integrated planner that models joint motion and grasp planning as an optimization problem. The computation of feasible and reachable grasps and the search for collision-free trajectories are connected.
    \item We propose an online learning algorithm and a grasp synthesis approach to select grasps and generate grasps online, which eliminates the need for a perfect grasp set and grasp selection heuristics.
    \item We showcase our algorithm for manipulation planning tasks in cluttered scenes. By comparing the proposed approach with existing methods, we demonstrate that our integrated planner improves performance.
\end{enumerate}

\section{Related Work}

\subsection{Motion Planning}
\subsubsection{Sampling-based Methods} 

Sampling-based motion planners such as Rapidly exploring Random Trees (RRTs) \cite{lavalle1998rapidly,kuffner2000rrt,karaman2011sampling} and Fast Marching Tree (FMT) \cite{janson2015fast} find trajectories by incrementally building space filling trees through directed sampling. RRTs offer probabilistic completeness. If there exists a solution to the problem, given sufficient time, a feasible trajectory can be found. However, sampling-based algorithms can be difficult to use in some applications due to the computational challenges and requirements for post-processing steps such as smoothing.

\subsubsection{Trajectory Optimization}

Trajectory optimization starts with a possibly infeasible trajectory and then optimizes the trajectory by minimizing a cost function. CHOMP \cite{ratliff2009chomp} and related methods \cite{dragan2011manipulation,dragan2011learning} optimize a cost functional using covariant gradient descent, while STOMP \cite{kalakrishnan2011stomp} uses stochastic sampling of noisy trajectories to optimize non-differentiable costs. More recently, TrajOpt \cite{schulman2014motion} solves a sequential quadratic program and performs convex continuous-time collision checking. GPMP2 \cite{mukadam2018continuous} formulates the problem as inference on a factor graph and finds the maximum a posteriori trajectory by solving a nonlinear least squares problem. Trajectory optimization methods are fast, but can only find locally optimal solutions. The local nature of previous methods can often be improved by multiple initialization and goals \cite{schulman2014motion,dragan2011learning}.

Our method is based on CHOMP~\cite{ratliff2009chomp} for trajectory optimization. Instead of optimizing trajectories on fixed endpoints, we also consider goal selection and synthesis in our planner.
%  Our trajectory interpolations to goal set have similar flavors as random trajectory samples in STOMP and mean prior trajectory in GPMP2.
 While \cite{dragan2011learning} model the goal selection problem as a regression on trajectory attributes through offline training, our method selects the target goal online during planning. 

\subsection{Grasp Planning} 

\subsubsection{Analytic Methods} 
Analytic methods \cite{nguyen1988constructing,ferrari1992planning,chen1993finding} often require full object information to define grasp qualities such as maximal disturbance resistance, force closure and antipodal grasps. For instance, \cite{ciocarlie2007dexterous} strive to reduce the dimensionality of the search space for robust grasps and use eigengrasps in the Graspit! simulator \cite{miller2004graspit} to plan with the simulated annealing algorithm. Recently, \cite{fan2018grasp} propose to find robust grasp poses by iterative surface fitting that optimizes error metrics of normal alignments and distances as a least squares problem.

\begin{figure*}
 \centering 
 \includegraphics[width=0.9\textwidth ]{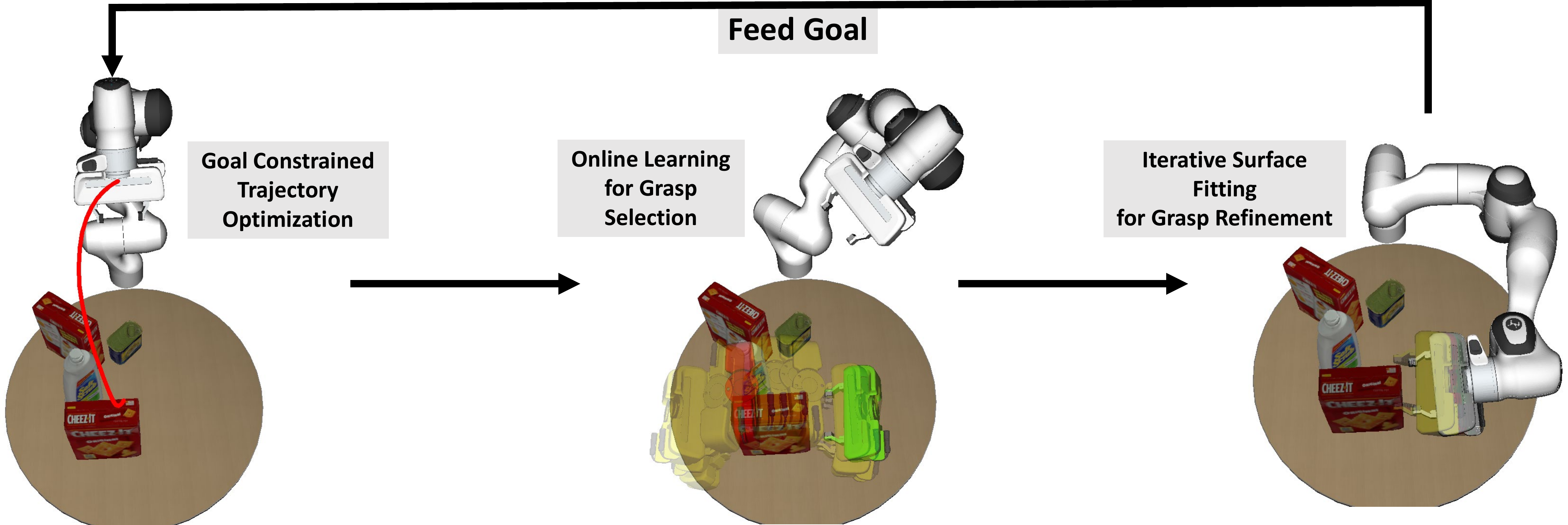}
   
   \caption{Our Optimization-based Motion and Grasp Planner (OMG Planner) consists of three components. The motion planner component optimizes the current trajectory denoted by the red line for a fixed target goal. The online learning component considers the goal set (yellow), and update the initial grasp (red) with a new grasp (green). The grasp refinement component iteratively improves the grasp. The refined grasp is used as the target goal for the next iteration.}
   \label{fig:overview}
   \vspace{-2mm}
\end{figure*}

\subsubsection{Data-driven Methods} 
Compared to analytic methods, data-driven approaches  \cite{goldfeder2011data,bohg2013data} have gained popularity since they can work with partial information of objects. For determining object grasps from images, \cite{redmon2015real} assume objects are put on a table top, and a parallel gripper grasps objects from top-down directions. Due to the multi-modality of grasping tasks, \cite{ten2017grasp} sample 6 DOF grasps and train a neural network to predict a quality score. \cite{mousavian20196} also focus on the diversity of grasp synthesis, which allows and benefits motion planning with multiple goals. Compared to these grasp planning methods, our method jointly plans arm motion and grasps to reach objects.

\subsection{Integrated Motion and Grasp Planning}

The advantage of an integrated planner is that motion planning and grasp planning impose constraints on each other. The closest works to ours are the integrated motion and grasp planners \cite{vahrenkamp2010integrated,fontanals2014integrated,haustein2017integrating}. These methods are built upon sampling-based motion and grasp planners. They bias a motion planner to sample nodes that lead to better grasps while a grasp sampling planner synthesizes grasps online. Different from these methods, we focus on studying the grasp selection problem in a trajectory optimization framework. Therefore, our method can utilize high-quality grasps generated from well-designed grasp planners, such as grasps synthesized from physics simulation \cite{miller2004graspit,clemen2019}.

\subsection{Task and Motion Planning}

TAMP methods focus on how to combine high-level task planning with low-level motion planning \cite{lagriffoul2014efficiently,srivastava2014combined,garrett2018sampling,toussaint2018differentiable}, where the tasks usually involve hybrid states and require multiple steps of actions such as picking, placing and pushing. For example, \cite{lagriffoul2014efficiently} leverage geometric constraints from motion planning to reduce the search space of possible actions. \cite{garrett2018sampling} build factored transition systems with sampling-based algorithms for robotic manipulation problems.  \cite{toussaint2018differentiable} formulate a constrained path optimization problem for sequential manipulation that involves physical interactions by logic geometric programming \cite{toussaint2015logic}.
Our approach is related to TAMP for manipulation planning. However, we neither perform any high-level reasoning nor model sequences of actions.
We focus on obstacle avoidance with kinematics constraints for grasping. While TAMP methods usually simplify or approximate the geometry of the manipulated objects, we do not make such approximations. Our method can benefit TAMP by providing a joint motion and grasp planner for task-level reasoning.

% In control literature, synthesis of movement that involves hybrid contact processes is typically modelled as mixed integer programming problems \cite{deits2014footstep}.

\subsection{Online Learning}

Compared with statistical learning in which data is assumed to come from a distribution and the goal is to minimize an excess risk, online learning instead considers a setting where a learner makes a decision that triggers a loss function online at every time step and the learner needs to dynamically improves its decision making by minimizing the accumulated losses \cite{bubeck2012regret,hazan2016introduction,lattimore2018bandit}. In the regret-based framework, the learner tries to minimize the regret on the loss sequence. Online interactive learning has been successfully applied to tackle different problems in robotics such as manipulation \cite{kroemer2010combining}, navigation \cite{godoy2015adaptive}, and model predictive control \cite{wagener2019online}.  Here, we formulate the grasp selection problem as an adversarial linear bandit \cite{lattimore2018bandit} in the online learning framework, where our learner (the planner) needs to choose the grasp goal online and adapt to some measured costs during every iteration of the trajectory optimization.

% A specific example is the Metrical Task Systems (MTS) problem \cite{borodin1992optimal}. In MTS, there are a service cost that depends on the environment whose behavior is hard to predict and a movement cost that penalizes large distribution shifts. In our goal selection problem, we do not have an accurate estimate of our costs and we can suffer from switching goals too often. So we model goal selection as an online bandit problem and apply online convex optimization approaches to the integration of motion and grasp planning. 

% Adaptive control is another related field. For instance, \cite{abbasi2011regret} derive a regret bound for a linear quadratic systems.

%In particular, an algorithm that has the sublinear regret bound in $T$ is nontrivially adapting to the incoming costs.

% Online learning algorithms often use additional regularization to improve stability.

\section{Methodology}

The key idea of our method is to simultaneously optimize the arm motion trajectory and the end-effector configuration for grasping. We formulate the problem as a constrained trajectory optimization problem, where we require the trajectory to avoid obstacles and to be smooth. Meanwhile, the last configuration of the trajectory should afford grasping. To achieve this, our method synthesizes and selects grasps online during trajectory optimization. Fig.~\ref{fig:overview} presents an overview of our optimization-based motion and grasp planner.

% The information about goal set informs our optimization to find better trajectory, while the motion generation guides the goal selection. Moreover, we iteratively refine the grasp quality of our synthesized goal set.

Formally, we define a trajectory $\xi: [0, 1] \mapsto \mathcal{Q}$ to be a function that maps time $[0,1]$ to the robot configuration space $ \mathcal{Q}$, where $\mathcal{Q} \subset \mathbb{R}^d$ and $d$ is the degree of freedom of the robot manipulator. Given a scene of multiple objects, we define the feasible grasp set of a target object as $\mathcal{G} \subset\mathcal{Q}$. Then we solve the following optimization problem to find the optimal trajectory of the robot manipulator for grasping:
\begin{align} \label{eq:optimization}
\xi^*= &\argmin \limits_{\xi} f_{\text{motion}}(\xi) \nonumber \\
&\text{s.t. } \xi(1)\in \mathcal{G},
\end{align}
where $f_{\text{motion}}$ is the objective function of the trajectory, and $\xi(1)$ indicates the last configuration of the trajectory. The constraint in Eq.~\eqref{eq:optimization} requires the last configuration at time $1$ to be a feasible grasping configuration. 

%We denote the optimization step with index $i$ and trajectory time with $t$. Our goal is to output a collision-free, smooth trajectory $\xi^*$  that ends at feasible grasp with finite horizon $N$. Thus, our objective function is formulated as follows. 
%\begin{equation}
%f(\xi)= f_{\text{motion}}(\xi), \text{ such that %$\xi(1)\in \mathcal{G}^*$},
%\end{equation}
%where $f_{\text{motion}}$ affects the entire trajectory of the robot arm between time interval $[0, 1]$, and the constraint requires the last configuration at time $1$ to be a feasible grasping configuration. 
%Moreover, in practice, we do not obtain $\mathcal{G}^*$ and have a single end configuration, so we propose to synthesize $\mathcal{G}$ and select grasp online. 

\subsection{Trajectory Objective Functional} \label{sec:chomp_obj}

Similar to CHOMP \cite{ratliff2009chomp}, we model the objective functional
\begin{equation} \label{eq:motion}
f_{\text{motion}}(\xi) = f_{\text{obstacle}}(\xi) + \lambda f_{\text{prior}}(\xi),
\end{equation}
where the obstacle term $f_{\text{obstacle}}$ bends the trajectory away from obstacles by penalizing parts of the robot that are close to or already in collision with objects in the scene, the prior term $f_{\text{prior}}$ measures the dynamical quantities across the trajectory such as velocities and accelerations, and $\lambda$ is a weight to balance the two terms. In this work, the prior term is defined as the integral over squared velocity norms:
\begin{equation}
f_{\text{prior}}(\xi) = \frac{1}{2}\int_0^1 \norm{\xi'(t)}^2dt,
\end{equation}
where $\xi'(t)$ indicates the velocity of the trajectory at time $t$.

To define the obstacle term, let $\mathcal{B} \subset \mathbb{R}^3$ be a set of body points on the robot
and $x(q, u): \mathcal{Q} \times \mathcal{B} \mapsto \mathbb{R}^3$ the forward kinematics mapping a body point $u \in \mathcal{B}$ with configuration $q \in \mathcal{Q}$ to the workspace.
Furthermore, let $c_{\text{obstacle}}:\mathbb{R}^3\mapsto \mathbb{R}$ be a workspace cost function that
penalizes points in the workspace inside and around the obstacles using the Signed Distance Field (SDF). Since we
want to drive the body points away from collision, the obstacle term in Eq.~\eqref{eq:motion} is an integral that collects the cost
of body points on the robot in the workspace along the trajectory:
\begin{equation}
f_{\text{obstacle}}(\xi)=\int_0^1\int_{\mathcal{B}}c_{\text{obstacle}} \big( x(\xi(t),u) \big) \norm{\frac{d }{dt}x(\xi(t),u)}dudt. 
\end{equation}
Please refer to \cite{ratliff2009chomp} for the definition of $c_{\text{obstacle}}$ and the derivation of the gradients of $f_{\text{prior}}$ and $f_{\text{obstacle}}$.

\subsection{Iterative Update Rule}
\label{sec:update_rule}
In practice, we use a discretization of the trajectory function over time: $\xi \approx (q_1^\top,q_2^\top,...,q_n^\top)^\top \in \mathbb{R}^{n\times d}$ for resolution $n$. Under this parametrization,
we can write the prior term as: \begin{equation}
f_{\text{prior}}(\xi)= \frac{1}{2}\norm{K\xi + e}^2=\frac{1}{2}\xi^\top A \xi + \xi^\top a +c,
\end{equation}
where $K$ is a finite differencing matrix, $e$ is a vector that accounts for the contributions of the configurations that remain constant in the trajectory, i.e., the start configuration and the
goal configuration, and $A=K^\top K$ is the dynamic matrix. The covariant idea of CHOMP comes from the inverse of the dynamic matrix. $A^{-1}$ acts as a smoothing operator that propagates the Euclidean gradient of the objective along the trajectory and seeks to make small changes in the average acceleration. Given this parametrization, CHOMP is a variant of
gradient descent that minimizes the linear approximation of $f_{\text{motion}}$ about $ \xi $ within an ellipsoid trust region. The distance metric that shapes this ellipsoid is defined by $A$, and we obtain the update rule in gradient descent for the $i$th iteration: \begin{equation} \xi_{i+1}=\xi_i - \frac{1}{\eta_{\text{motion}}}A^{-1}v_i ,\end{equation} where $v_i=\nabla f_{\text{motion}}(\xi_i)$ is the discretized functional gradient and $\eta_{\text{motion}}$ is the step size.

\subsection{Goal Set Constraint}

The end-point of $\xi_i$ can be varied during planning for a set of feasible goal configurations $\mathcal{G}$. Following \cite{dragan2011manipulation}, this is achieved by introducing a constraint on the trajectory. Given a general constraint $h(\xi)=0$, we can linearize $h$ around $\xi_i$: $h(\xi)\approx h(\xi_i)+\frac{\partial}{\partial \xi}h(\xi_i)(\xi - \xi_i)=C(\xi -\xi_i) +b$, where $C$ is the Jacobian of $h$ evaluated at $\xi_i$ and $b=h(\xi_i)$. We use Lagrangian to solve the linearized constraint and yield the projected gradient update rule: 
\begin{align}
\xi_{i+1}= \xi_i &-\frac{1}{\eta_{\text{motion}}}A^{-1}v_i\nonumber \\&+ \frac{1}{\eta_{\text{motion}}}A^{-1}C^\top(CA^{-1}C^\top)^{-1}CA^{-1}v_i \nonumber \\
&-A^{-1}C^\top(CA^{-1}C^\top)^{-1}b.
\label{eq:chomp_proj}
\end{align}
Assume our goal configuration is $g$ at the $i$th iteration. We simply model the goal set constraint as
\begin{equation} \label{eq:chomp_proj_constraint}
h(\xi_i) = \xi_i(1)-g = 0,
\end{equation}
and denote the update rule in Eq.~\eqref{eq:chomp_proj} by \textit{CHOMP-Proj}($\xi,g$). 
With this update rule, we denote $f_g(\xi_i)$ as the objective cost if we use $g$ as the goal from the $i$th iteration. As we move the end-point, CHOMP smoothly updates the remaining trajectory. This formulation is different from the vanilla CHOMP in that the goal is not fixed and can change in different iterations.

\subsection{Online Learning for Grasp Selection}
\label{sec:online_learning}
While \cite{dragan2011manipulation} address the problem of CHOMP planning with a goal set constraint, modeling goal selection as projection by Eq.~\eqref{eq:chomp_proj_constraint} can often be simplistic. \cite{dragan2011learning} propose a few trajectory attributes such as goal radius, elbow room, distance, etc., for a regression formulation. Indeed, the optimality of a goal should be induced by the objective function and thus related to the success of motion planning. Given that we can optimize $f_g(\xi_i)$ with a target goal, we propose to choose the goal that maximizes our motion generation success under an online linear bandit scheme \cite{bubeck2012regret,lattimore2018bandit}. An online linear optimization framework is suitable because at every iteration, we can model the objective costs of the goal set as our suffered losses and relax our goal selection as probability distributions.

Assuming the goals in $\mathcal{G}$ are intricately correlated, our strategy should dynamically choose the goal for the next iteration until it finds the best goal. In CHOMP projection, we suffer from switching to goals that can heavily perturb our current trajectory into collision. Therefore, to allow for principled behavior that selects the next goal $g_{i+1}$, we need a strategy that balances exploitation and exploration of the costs.

In the $i$th iteration, denote the goal distribution on a discrete goal set $\mathcal{G}$ by $p_{i}$. We apply online learning to solve for $p_{i+1}$. To use optimization for goal selection, we need to define the cost of each goal in the goal set, denoted by a cost vector $c_i$. Ideally, we would like to have $c_i(g)=f_{g}(\xi_i)$, then the objective cost directly tells us which target goal to use. However, since $f_{g}(\xi_i)$ is not available without running CHOMP for each goal, we generate surrogates to our objective cost $f(\widehat{\xi}_i^g) \approx f_{g}({\xi}_i)$ by estimating the potentially optimized trajectory $\widehat{\xi}_i^g$ as if the trajectory $\xi_i$ ends at $g \in \mathcal{G}$.

\begin{figure}
 \centering 
 % \includepdf[]{flow_chart.pdf}
 \includegraphics[width=0.48\textwidth]{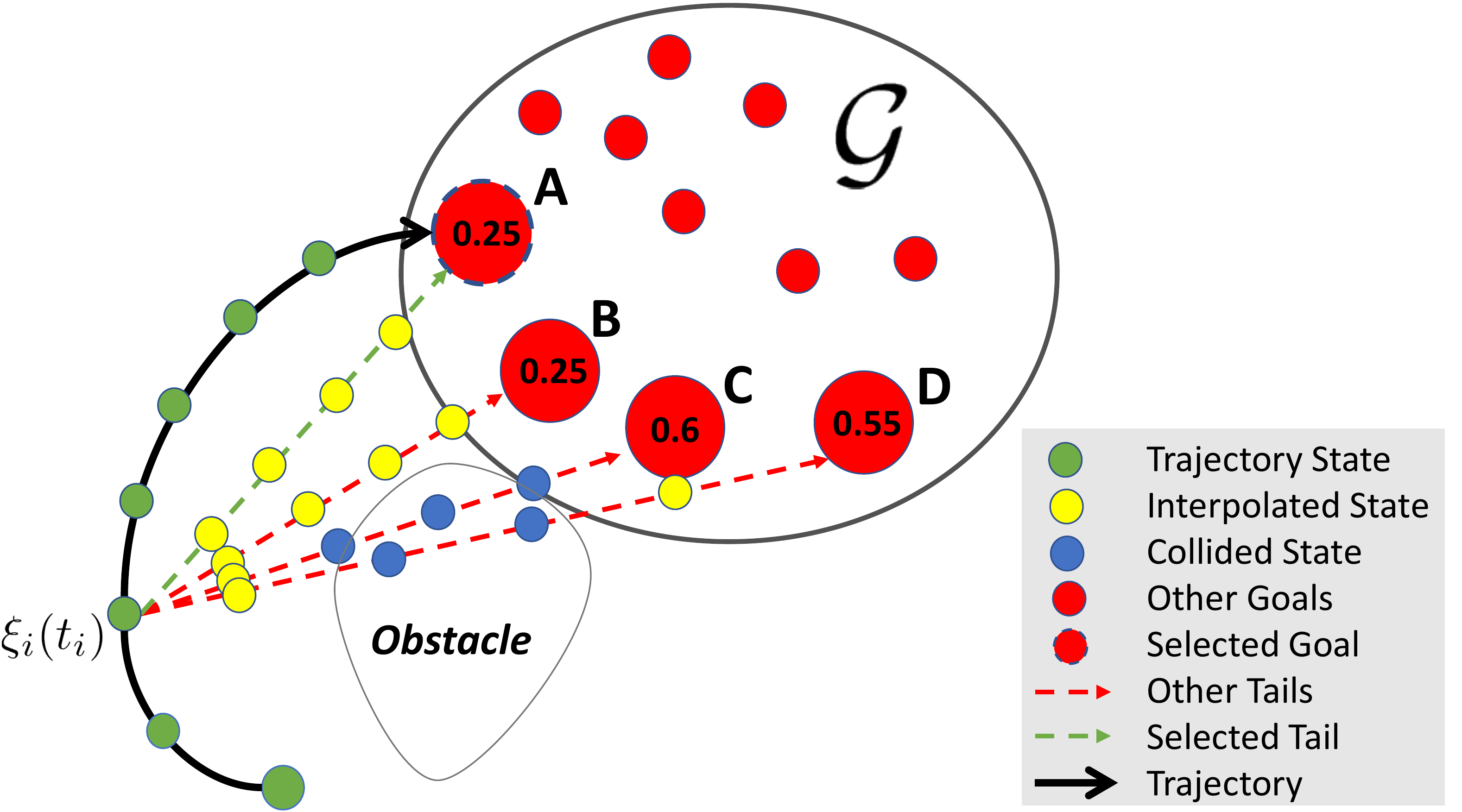}

   \caption{Online goal selection algorithm at the $i$th iteration. The goal costs $c(g)$ are denoted inside red circles. In this example, the online learning algorithm is about to shift goal from $A$ to $B$, which yields a collision-free yet shorter trajectory.}
    \label{ol}
    \vspace{-3mm}
\end{figure}

We propose that to compute $f(\widehat{\xi}_i^g)$, it is more natural to look at the tail of trajectory denoted by $\widetilde{\xi}_i^g$ starting at $t_i =\frac{i}{N}$, where $N$ is our optimization horizon. To motivate this, recall that in each \textit{CHOMP-Proj} step, we project onto the goal constraint and propagate the changes smoothly. Therefore, as the time step $i$ approaches $N$, the changes to the trajectory impact more on the tail than the head of the trajectory. Our estimated costs would be more accurate if we use the tail $\widetilde{\xi}_i^g$ instead of the whole trajectory. Moreover, the goal cost vector needs to converge for stability of the algorithm and it should also be dynamic from the learning perspective. Another way to think about this is that we are ``simulating'' the optimal control execution of the trajectory during planning and estimating the running cost for dynamic goal selection. As shown in Fig.~\ref{ol}, since the prior of CHOMP is the minimum smoothness trajectory, one can approximate the trajectory tail $\widetilde{\xi}_i^g$ as the interpolation between $\xi_i(t_i)$ and the goal $g$ with constant velocity. The associated cost for the next iteration is defined as $c_{i+1}(g)=f(\widetilde{\xi}_i^g)$ and normalized to have unit norm. 

Given our estimated objective cost $c_{i+1}=\{f(\widetilde{\xi}_i^g)\mid g\in \mathcal{G}\}$ and previous costs $c_1,...,c_i$ for the goal set $\mathcal{G}$, we apply online learning algorithms to update the goal distribution $p_{i+1}$ as described later. 
Then the expected loss at the $(i+1)$th iteration under this distribution is $\langle c_{i+1}, p_{i+1} \rangle $ and $g_{i+1}$ is defined as the mode of the distribution $p_{i+1}$, which turns out to be more stable than a sample. Overall, online learning for goal selection aims to minimize the regret
\begin{equation} \label{eq:regret}
  \mathcal{R}_N=\sum\limits_{i=1}^N \langle c_i, p_i\rangle -\min\limits_{g \in \mathcal{G}} \sum\limits_{i=1}^N c_i(g),
\end{equation}
which compares the loss of our strategy $p_i$ up to time $N$ with the best cost of any fixed goal in hindsight. In the following, a few commonly used optimization methods for online learning problems \cite{bubeck2012regret} are presented. These methods can be used to solve $p_{i+1}$ at the $i$th iteration.

\subsubsection{Follow the Cheapest (FTC)} FTC greedily moves the distribution $p_{i+1}$ entirely to the point that has the minimum cost $c_{i+1}(g)$. \textit{CHOMP-Proj} turns out to be a FTC algorithm with the costs defined by the distances between the end-point of the current trajectory and the goals. 

\subsubsection{Follow the Leader (FTL)} FTL computes the point $g \in \mathcal{G}$ that causes the lowest total cost $\sum_{j=1}^{i+1} c_j(g)$ and then makes the distribution $p_{i+1}$ concentrated entirely on $g$.  

\subsubsection{Exponential Weighting (EXP)} EXP begins with an uniform distribution over goals: $p_1(g)=\frac{1}{|\mathcal{G}|}, \forall g \in \mathcal{G}$. At time $i$, it first multiplies the previous probabilities with the corresponding exponential costs: $\widehat{p}_{i+1}(g)=\exp(-\eta_{\text{ol}} c_{i+1}(g))p_{i}(g), \forall g \in \mathcal{G}$, where $\eta_{\text{ol}}$ is the learning rate. Then it normalizes the vector to make it a probability distribution: $p_{i+1}(g)=\widehat{p}_{i+1}(g)/{\sum_{g\in \mathcal{G}}\widehat{p}_{i+1}(g)}$.

\subsubsection{Mirror Descent (MD)} We refer the readers to \cite{beck2017first,lattimore2018bandit} for this form of regularized optimization in a bandit setting. To allow MD to prefer probability distributions that are more diversified, we associate the Bregman divergence with entropy, which becomes the KL divergence. At time step $i$, the MD update rule is $p_{i+1}=\argmin\limits_{p\in \Delta} \eta_{\text{ol}} \langle   c_{i+1}, p  \rangle +D_{KL}(p ||p_{i})$, where $\Delta$ is the probability simplex and $\eta_{\text{ol}}$ is the learning rate. The $\argmin$ of this update rule can be solved by Lagrangian.

 \begin{figure}
\centering
 \includegraphics[width=0.35\textwidth]{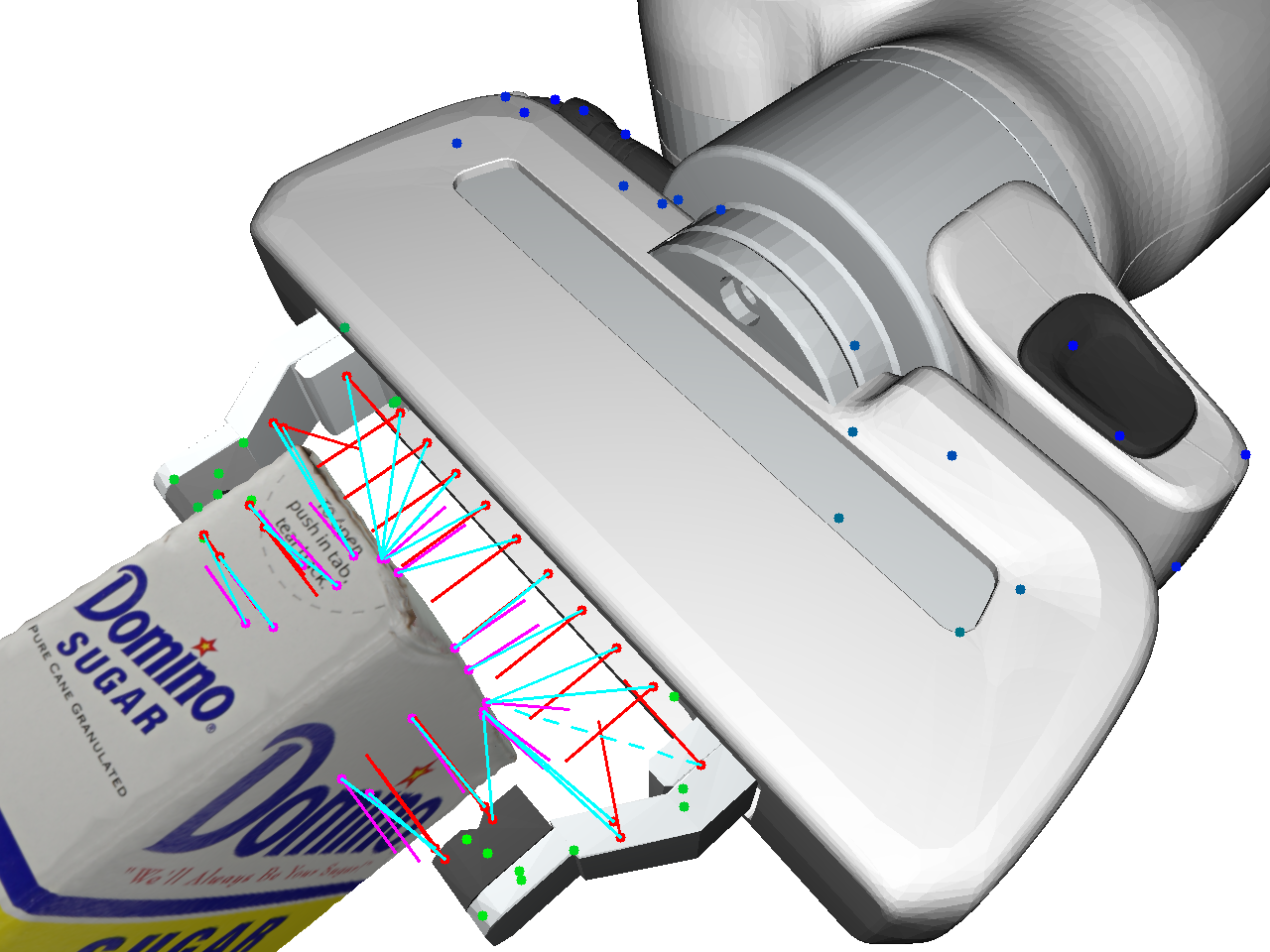}
\caption{Red points denote the sampled contact points $u_h$ with normals $u_h^n$ and purple points denote the corresponding object points $u_o$ with normals $u_o^n$. The correspondence lines are colored in cyan. Body points $u$ range from blue to green, denoting from far to close with respect to objects.}
\label{fig:hand_points}
\vspace{-4mm}
\end{figure}

 \subsection{Online Grasp Synthesis}
 \label{sec:grasp_synthesis}
So far we have assumed that the goal set is given and fixed. A fixed set of grasps limits the number of grasp possibilities. Even with a set of perfect grasps in the workspace, it is unclear that solving inverse kinematics would give a sufficient goal set in the configuration space. A fixed set also has low adaptability to the environment since it fails to explore new grasps. 

Starting from an initial finite goal set $\mathcal{G}$, we propose to synthesize more candidate grasp configurations online. Suppose after the online learning process for goal selection, the configuration end goal $g_i$ is selected at iteration $i$. For simplicity, we drop the iteration index $i$ in the followings. Our proposed grasp synthesis algorithm aims to improve the grasp affordance by minimizing the objective function:
 \begin{equation}
    f_{\text{grasp}}(g)=f_{\text{isf}}(g)+  \gamma f_{\text{collision}}(g),
\label{eq:grasp}
\end{equation} 
where $f_{\text{isf}}(g)$ measures the grasp quality using the Iterative Surface Fitting (ISF) algorithm, $f_{\text{collision}}(g)$ measures collisions in the scene, and $\gamma $ is a constant to balance the two terms. We use $f_{\text{isf}}(g)$ to refine the end configuration $g$ such that the end effector transformation $T \in \mathbb{S}\mathbb{E}(3)$ makes the surface of the robot gripper $\mathcal{S}$ match against the surface of the target object $\mathcal{O}$. We softly penalize hand and arm collisions and use gradient descent to optimize the objective. This enables efficient and stable optimization in the configuration space.

Specifically, we generalize the forward kinematics mapping $x(g,u)$ of a body position $u$ in Sec.~\ref{sec:chomp_obj} to include directions: $x(g,u^n)$ maps a body direction vector $u^n$ and a robot configuration $g$ to a direction vector in the workspace. Observe that the end effector transformation $T$ in the workspace is the hand coordinate transformed by $g$. We can write all workspace costs on $T$ as the configuration space costs on $g$. We define a set of hand contact points $\{h_j\}$ and a set of contact normals $\{h_j^n\}$ on the robot hand and fingers, indexed by $j$ (see Fig.~\ref{fig:hand_points}). Denote $u_h=\{x(g,h_j)\}$ and $u_h^n=\{x(g,h_j^n)\}$ as the two sets in the workspace transformed by forward  kinematics. We then search for the nearest neighbor point $u_o$ and normal $u_o^n$ on object $\mathcal{O}$ associated with $u_h, u_h^n$. For $m$ pairs of points and normals, we define the point matching loss as
\begin{equation}f_{\text{pml}}(g)=\sum \limits_{j=1}^m \langle u_{h,j}-u_{o,j}, u_{o,j}^n \rangle^2,
\end{equation} 
and the normal alignment loss as
\begin{equation}f_{\text{align}}(g)=\sum \limits_{j=1}^m (\langle u_{h,j}^n, u_{o,j}^n \rangle + 1)^2. 
\end{equation} 
The point matching loss closely attaches the hand to the object, and the normal alignment loss aligns their surface normals. Overall, the ISF loss is defined as 
\begin{equation}f_{\text{isf}}(g)=f_{\text{pml}}(g)+\alpha f_{\text{align}}(g),
\end{equation} where $\alpha$ is a weight coefficient.

Since the hand has to remain collision-free with respect to the object, we evaluate the object points in the SDF of the robot hand and penalize any hand collision $f_{\text{hand\_obstacle}}(g)$. This is different from the obstacle cost in CHOMP, where robot body points are evaluated in the SDFs of objects. Note that the gradient for object points in the SDF of the hand can be treated as hand points pushing in opposite gradient directions. We also add the CHOMP obstacle cost evaluated at a single configuration $g$, $f_{\text{obstacle}}(g)$, to allow local arm obstacle avoidance. The overall collision cost for grasp synthesis is  
\begin{equation}f_{\text{collision}}(g)=f_{\text{hand\_obstacle}}(g)+\beta f_{\text{obstacle}}(g), \end{equation}
where $\beta$ is a balancing weight.

Algorithm 1 presents our C-Space ISF algorithm for grasp refinement. The workspace cost $f_{\text{grasp}}$ in Eq.~\eqref{eq:grasp} and its gradient $\nabla f_{\text{grasp}}$ are computed by the $Error$ function. The gradient $\nabla f_{\text{grasp}}$ is updated via the exponential map $exp$ from $\mathit{s}\mathit{e}(3)$ to $\mathbb{S}\mathbb{E}(3)$ and pulled back to the configuration space $\nabla g$ through the Jacobian transpose $J^\top$ with a step size $\eta_{\text{grasp}}$. We then update the grasp $g$ in $\mathcal{G}$ and return back to motion planning step. In practice, one can run the C-Space ISF algorithm \ref{Algorithm1} multiple times for each motion planning iteration.

\begin{algorithm}

\caption{C-Space ISF $(g_{i}, u_h, u_h^n, \mathcal{O}, \mathcal{S})$}
\label{Algorithm1}
    $u_o,u_o^n=\textit{NearestNeighbor}(u_h,u_h^n, \partial \mathcal{O})$\\
  $ f_{\text{grasp}}, \nabla f_{\text{grasp}}=  \textit{Error}(u_h,u_h^n, u_o,u_o^n, \mathcal{O}, \mathcal{S})  $ \\ 
  $ \nabla g_i =  J^\top\textit{exp}(\nabla f_{\text{grasp}})$ \\
   $  g_i = g_i - \eta_{\text{grasp}}  \nabla g_i$
\end{algorithm}

\subsection{Joint Motion and Grasp Planning}

We have defined our iterative update rule for the $i$th iteration, involving optimizations of smoothness, collision, goal selection, and grasp quality. Our integrated planner simply runs this iterative update rule for $N$ times. A pre-termination criterion can be set with the threshold on grasp quality, trajectory smoothness, and collision-free requirement. Our joint planning algorithm is shown in Algorithm \ref{Algorithm3}. \textit{CHOMP-Proj} provides a way to smoothly update the trajectory to avoid obstacles given a goal. The \textit{ONLINE-LEARNING} as described in Sec.~\ref{sec:online_learning} considers all grasps in the goal set and updates our goal distribution for the next iteration. The \textit{C-Space ISF} applies to the selected goal to refine its quality.

It is worth to mention that we tried to formulate a joint optimization problem by combining the two objective functions in Eq.~\eqref{eq:optimization} and Eq.~\eqref{eq:grasp}. However, we found that the ISF algorithm heavily depends on the initialization and is often stuck in local optima with bad initializations. Our current formulation provides good initialization to the ISF algorithm using the goal set, which makes the optimization more stable.

 \begin{algorithm}

\SetAlgoLined
\caption{OMG Planner}
\label{Algorithm3}
 initial trajectory $\xi_0, \mathcal{O}, \mathcal{G}, \mathcal{S}, p_0, g_0$ \\
\For{ $i = 0,...,N$}{
   $\xi_{i+1}=\textit{CHOMP-Proj}(\xi_i,g_i)$\\
    
    $p_{i+1} = \textit{ONLINE-LEARNING}(\xi_{i+1}, \mathcal{G})$ \\
    $g_{i+1} = \argmax({p_{i+1}})$ \\
    $g_{i+1}=\textit{C-Space ISF}(g_{i+1}, \mathcal{O}, \mathcal{S})$\\
     }
    
\end{algorithm}

\section{Experiments}

We conduct experiments with the Panda Franka arm, a $7$-DOF arm with a parallel gripper (Fig.~\ref{fig:overview}). We generate cluttered scenes by dropping sampled YCB objects \cite{calli2015benchmarking} with random upright poses on a table in the Pybullet Simulator \cite{coumans2013bullet}. Each scene samples 3 to 7 objects. The objects used are master chef can, cracker box, sugar box, tomato soup can, mustard bottle, potted meat can, pitcher base, bleach cleanser, bowl, and mug. 

\begin{figure}
\centering
 \includegraphics[width=0.4\textwidth]{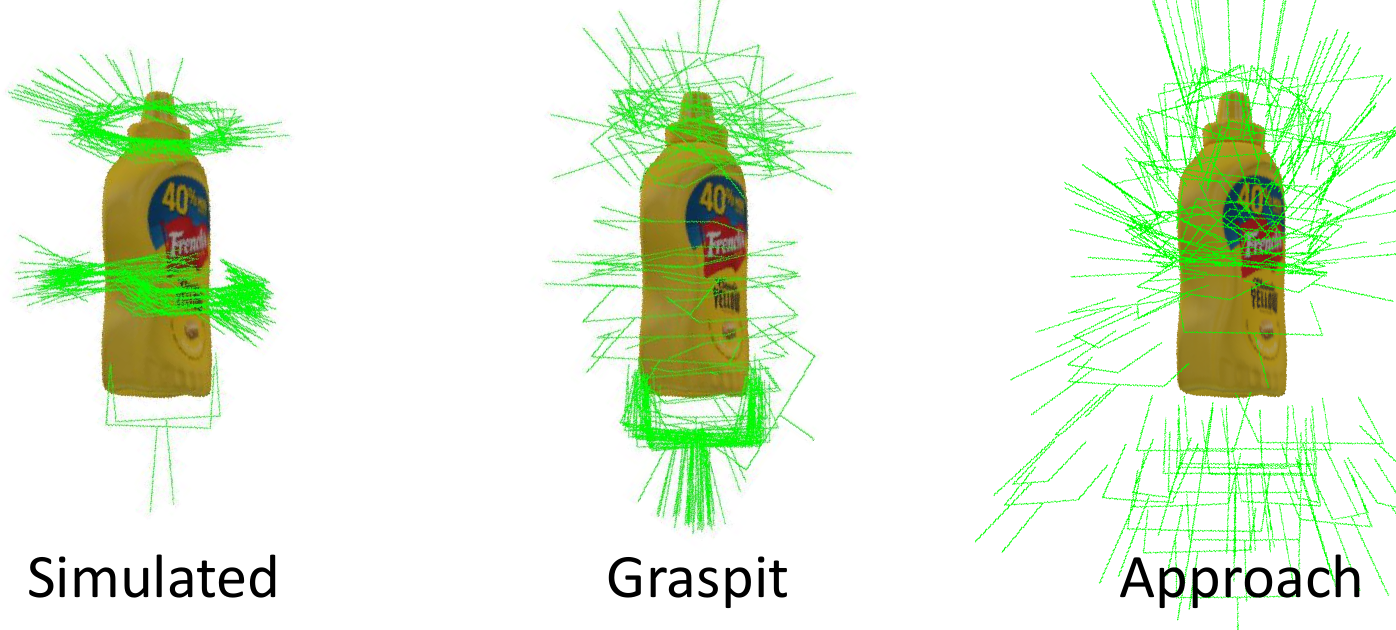}
\caption{Three initial grasp sets of the mustard bottle.}
\label{fig:grasp_sample}
\vspace{-4mm}
\end{figure}

We experiment with three different sets of initial grasps as shown in Fig.~\ref{fig:grasp_sample}. The ``Simulated'' grasps are sampled from a physics-based simulator \cite{clemen2019}. The second set is sampled from the Graspit! planner \cite{ciocarlie2007dexterous}. The ``Approach'' grasps are collision-free grasps naively sampled by random approach directions. The initial start configuration is fixed, and goal configurations are generated from inverse kinematics and checked collision-free. A trajectory is then initialized as a cubic spline connecting the start configuration and the initial goal.

\subsection{Evaluation Metrics}

Motion planning methods are usually evaluated by the feasibility of the generated paths, i.e., collision-free paths. In addition, we measure several attributes of the paths that are desirable in our grasping tasks. 

Trajectories close to obstacles should be avoided in practice. We measure ``collision'' as the sum of the obstacle costs $c_{\text{obstacle}}$ of the sampled points inside the minimum padding distance (5cm) by densely interpolating each trajectory to contain $200$ time steps. The smoothness of a trajectory is another desired property. We measure ``smoothness'' by the squared velocity norm as in $f_{\text{prior}}$. To measure the planning performance, we define the ``planning'' success as a collision-free path with its smoothness below a threshold of $30$.

% While path length is often used as a trajectory quality for fixed end-point settings, it is less applicable for a goal set algorithm as certain good goals away from obstacles might be farther to reach.

After planning, it is important to see how the plan is executed for grasping. The success of grasp execution involves system dynamics, feedback controllers and grasp robustness, in addition to trajectory collision and smoothness. We execute a plan in the Pybullet simulator and define the ``execution'' success if the target object is lifted. The execution success rate takes failed plans into account as well. We also compute the ``grasp'' cost $f_{\text{grasp}}(\xi(1))$ from our C-Space ISF method to measure the grasp quality. Finally, planning ``time'' is measured in seconds for different algorithms.

\begin{table} \setlength{\tabcolsep}{5.7pt}
\small
\centering
\begin{tabular}{|c|c|c||c|c|c|c|}
 \hline
 \multicolumn{7}{|c|}{ 100 scenes, 30 grasps each scene, 10 runs} \\
\hline
 Algorithm & Fixed & Proj & FTC & FTL & EXP  & MD  \\
 \hline
 Execution&     74$\%$&  75$\%$&   \textbf{81$\%$} & 79$\%$&  80$\%$&  \textbf{81$\%$}\\
 Planning&     78$\%$&  79$\%$&   91$\%$& 89$\%$&  90$\%$&  \textbf{92$\%$}\\
 Smoothness&   18.87 &   \textbf{11.24}&   15.78& 15.86 &  15.71 &   15.54\\
 Collision &     5.97  &   12.06 &   4.68 & 4.10  &  4.95  &   \textbf{3.87} \\
 Time &      0.18  &   \textbf{0.15} &  0.20 & 0.22  &  0.23  &   0.30 \\
\hline

\end{tabular}
\caption{Comparison between two baselines and four online learning algorithms for grasp selection.}
\vspace{-2mm}
\label{tab:ol}
\end{table}

\subsection{Implementation Details}
 
\subsubsection{CHOMP}
We set the smooth weight $\lambda=0.1$, discretization resolution $n=30$, and learning rate $\eta_{\text{motion}}=0.01$. The obstacle padding is 0.2m within which the cost increases quadratically \cite{ratliff2009chomp}. A collision-free path is assumed to have a minimum distance of 5cm from obstacles. Instead of modeling the arm with swept spheres as in the original CHOMP, we uniformly sample body points from the vertices of the robot 3D mesh, and select $500$ points in the trajectory with the largest obstacle costs to update the obstacle gradient (see Fig. \ref{fig:hand_points}). These modifications make the integral over $\mathcal{B}$ more general and precise as well as stabilize the optimization. We sample 10 points from each body part in our implementation. Since grasping tasks require collision-free gripper poses, we also include sampled points from hand and fingers.

\subsubsection{Online Learning} The EXP algorithm uses the learning rate $\eta_{\text{ol}}=\sqrt{\log(|\mathcal{G}|)/N}$. The MD algorithm takes the learning rate over an ensemble of experts with $\eta_{\text{ol}}=2^k\log(N)$ and $k$ is selected from $\{-2,-1,0,2,4 \}$.
% In practice, we mix our distribution $p_t$ with a uniform distribution to accelerate adaptation to costs. 

\subsubsection{C-Space ISF}
We set the normal penalty weight $\alpha=0.01$, the arm collision coefficient $\beta=0.001$, the hand collision constant $\gamma=0.5$, and the learning rate $\eta_{\text{grasp}}=0.05$. The nearest neighbors are found with a KD-Tree on 1000 sampled points for each object, where repeated neighbors are filtered out to increase robustness.

% \subsection{Planning Results}

% In this section, We evaluate the algorithm on various experiment domains, and show that it achieves reasonable results in grasping tasks. In addition, we investigate the components that contribute to its success in ablation study. Finally, we show that this algorithm improves upon the planning success in terms of our objective and grasp quality.

% We investigate the following questions: 1) Does online learning help select goals in the grasp set to improve motion generation? 2) Can C-Space ISF refine grasp success?  3) Is there improvement on overall planning success by jointly optimizing grasp and motion? 4) In practice, what is the sensitivity of hyperparameters such as $n$ and $\lambda$ and their influence on smoothness and planning success? 

\subsection{Grasp Selection}

We first evaluate different algorithms for grasp selection in our trajectory optimization framework. In this experiment, the ``Simulated'' grasp set is used and no grasp refinement is performed. Therefore, the dominating factor in planning performance is the goal selection strategy. Table~\ref{tab:ol} presents the results for 100 scenes, where we compare the four online learning algorithms with two baselines. 1) ``Fixed'' selects the best goal based on the initial objective estimate and fixes the goal during planning. 2) ``Proj'' \cite{dragan2011manipulation} uses the \textit{CHOMP-Proj} update rule which always projects to the closest goal during trajectory optimization.

Our results in Table~\ref{tab:ol} indicate that the online learning methods improve the success of planning and execution. From the improvement over ``Fixed'', it is clear that planning toward a fixed goal does not provide the same performance as online algorithms that adaptively select goals, since the initial goal may be sub-optimal. Comparing the four online learning algorithms with ``Proj'', we can see that the cost estimate based on our objective function performs better than just using the distance between the trajectory end-point and grasps in the goal set. In addition, modeling the probability distribution of the goal set for grasp selection is beneficial. Our empirical analysis shows that these online learning methods provide a scheduling behavior for a planner to adaptively switch goals while keeping optimization stability. While the grasp set size presents a trade-off between runtime and performance, we find 30 grasps is a good balancing point. Finally, MD performs slightly better than the others with a trade-off in running time.

% Comparing ``Baseline'' with ``Proj'', we observe that a good initial goal simplifies the problem by a great extent. Comparing the four online learning algorithms with ``Proj'', we can see that the cost estimate based on our objective function performs better than just distance. Finally, from the improvement over ``Baseline'', it is clear that planning toward a fixed best goal does not provide the same performance as online algorithms that adaptively select goals. 

\begin{table}
\setlength{\tabcolsep}{1.4pt}
\small
\centering
\begin{tabular}{ |c|c|c|c|c|c|c| }
 \hline
 \multicolumn{7}{|c|}{100 scenes, 60 grasps, 30 iteration refinements, 10 runs} \\
 \hline
 Object Name & S & S Refined & G & G Refined & A & A Refined  \\
 \hline
Cracker Box &  2.47  &    \textbf{2.32} &   4.27 &    3.31 &  6.58 &  4.36 \\
 Mug &    5.11 &   \textbf{3.93} &  7.24  &  5.55 &   9.05  &   5.18 \\
 Tomato Soup Can   &   2.38    &  \textbf{2.27}  &  3.15  &    2.83  &   7.06  &    6.27 \\
 Mustard Bottle &   4.40   &   \textbf{3.19}   &    4.91  &     3.79   &   8.15  &   5.73 \\
 Potted Meat Can&   2.96  &   \textbf{1.95} &   3.08 &    2.54  &   7.47 &  4.91 \\
 Pitcher Base &  4.41  &   \textbf{2.97}  &   5.68  &   3.76  & 8.69  &   6.14  \\
 \hline
 Mean &   3.62  &   \textbf{2.77}  &   4.72  &   3.63  &   7.83 &  5.43 \\
 \hline
 \hline
 Planning & 92$\%$  &   \textbf{93\%}  & 84$\%$ & 84$\%$  &71$\%$ & 73$\%$ \\
 \hline
 Execution & 80$\%$  &   \textbf{82\%}  & 56$\%$ & 59$\%$  &26$\%$ & 33$\%$ \\
 \hline
\end{tabular} 
\caption{Grasp costs for six YCB objects before and after refinement. ``S'', ``G'' and ``A'' denotes the simulated set, the graspit set, and the approach set, respectively.}
 \label{tab:grasp_table}
\vspace{-2mm}
\end{table}

\begin{figure}
\centering
 \includegraphics[width=0.45\textwidth]{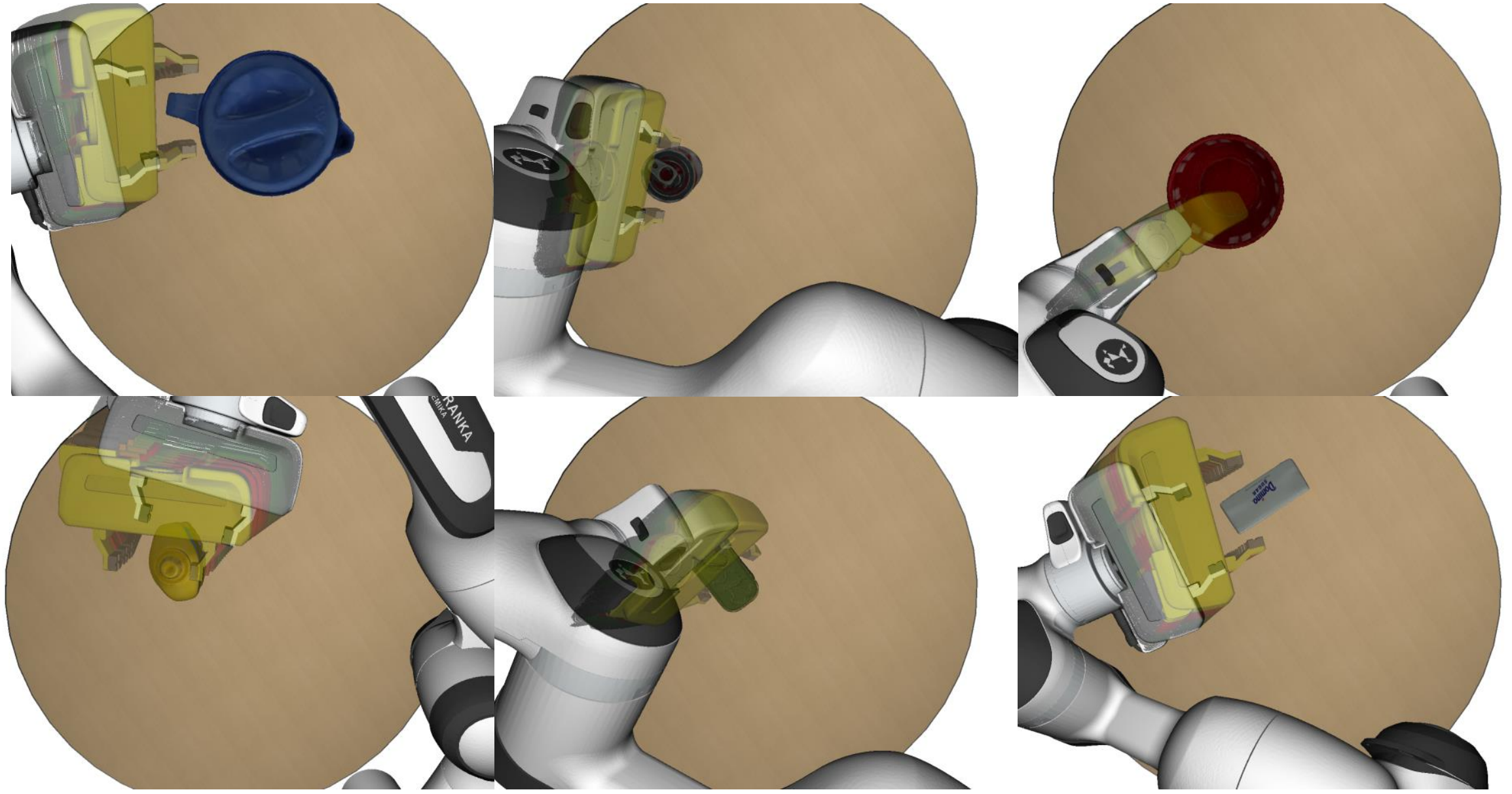}
\caption{Examples of grasps before and after refinement.}
\label{fig:graps_opt}
\vspace{-4mm}
\end{figure}

\begin{figure*}
 \centering 
 \includegraphics[width=0.95\textwidth]{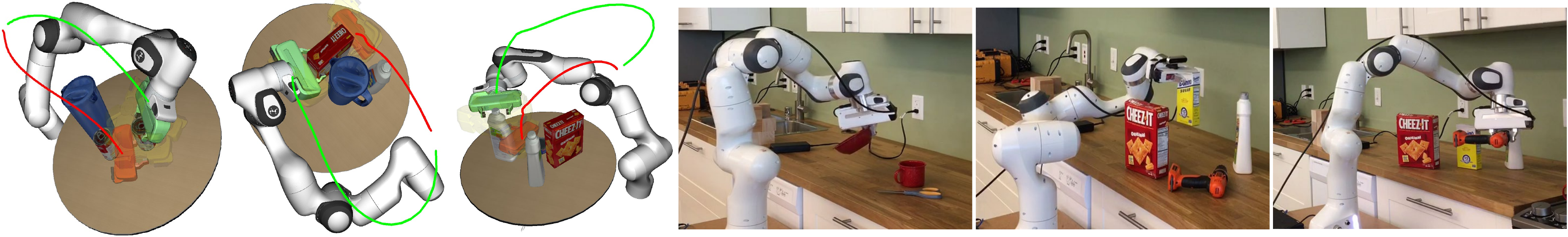}
\caption{Left) Planning scene examples with CHOMP trajectories in red and our OMG-Planner trajectories in green. Right) Successful real world grasping examples by executing trajectories from our planner using estimated 6D object poses from \cite{deng2020self}.}
\label{fig:planning_scenes}
\vspace{-2mm}
\end{figure*}

\subsection{Grasp Refinement}

We investigate the performance of the C-Space ISF algorithm for grasp refinement. In this experiment, we apply the MD algorithm for grasp selection and test the C-Space ISF algorithm on the three different grasp sets (Fig.~\ref{fig:grasp_sample}). Table~\ref{tab:grasp_table} shows the detailed grasp costs and the planning and execution success before and after grasp refinement.

% We test the grasp refinement performance on YCB Objects by running the C-Space ISF algorithm for 30 iterations. The optimization procedure minimizes the surface fitting error while avoiding the in-collision configurations. 

Overall, we observe a consistent improvement on different initial grasps from grasp refinement. The ``Simulated'' set achieves the minimum grasp cost and the highest success rate. Because these grasps are sampled and verified in simulation by grasping and shaking an object to see if the object falls or not \cite{clemen2019}. The drawback is that it is time-consuming to generate these high-quality grasps. Our refinement algorithm can further improve the quality of these simulated grasps. We can also see that the C-Space ISF algorithm heavily depends on the initial grasps. Therefore, the improvement on the ``Graspit'' set and the ``Apporach'' set is limited. This is also the reason why we combine grasp selection with grasp refinement, instead of optimizing grasps from random initial grasps. Fig.~\ref{fig:graps_opt} shows some qualitative examples of our grasp refinement method. 
 
% Without the availability of grasp databases such as ``Simulated'' and Graspit!, we can synthesize grasps as in ``Apporach'' efficiently and rely on refinement steps. For some fine grasps such as those clustered around the bottle neck, it would be hard to run our refinement. Also, the contact point association and the gradient update are affected by the initial grasp poses and hyperparameters, since they depend on local information.

\begin{table} \setlength{\tabcolsep}{0.01pt}
\small
\centering
\begin{tabular}{|c|c|c|c|c|c|}
 \hline
 \multicolumn{6}{|c|}{ 100 scenes, 30 grasps each scene, 10 runs} \\
\hline
 Algorithm &    CHOMP~\cite{ratliff2009chomp} & CHOMP-C~\cite{dragan2011manipulation} & FMT~\cite{janson2015fast}& RRT-C~\cite{kuffner2000rrt}  & OMG\\
\hline
 Execution&     77$\%$& 75$\%$&   66$\%$&  64$\%$& \textbf{84$\%$}\\
 Planning&      82$\%$&  79$\%$&    80$\%$&  78$\%$& \textbf{93$\%$}\\
 Smoothness&    19.12   & \textbf{11.24} & 27.21 &  26.45 &  15.43 \\
 Collision &     4.78 &  12.06&     6.37  &  6.52 & \textbf{3.92} \\
 Grasp &      3.43  &  3.35 &  3.49  &  3.48 & \textbf{3.04} \\
 Time & 0.25 & \textbf{0.15} & 0.67  &  0.22  & 0.42 \\
\hline
\end{tabular}
\caption{Comparison on planning methods for joint motion
and grasp planning.}
\label{tab:all_planning_results}
\vspace{-4mm}
\end{table}

\subsection{Planner Performance}

We compare our OMG planner with several other motion planners in Table~\ref{tab:all_planning_results}, where the ``Simulated'' grasp set is used. The OMG planner uses the MD algorithm for grasp selection and the C-Space ISF algorithm for grasp refinement. Since grasp planning is not covered by CHOMP~\cite{ratliff2009chomp}, FMT~\cite{janson2015fast} and RRT-Connect~\cite{kuffner2000rrt}. We adopt the common routine used in practice to generate manipulation trajectories for them: 1) ranking the goals in the grasp set based on heuristics such as our objective estimates; 2) planning through the goal set until a feasible trajectory is found. We perform planning experiments for 100 scenes with 30 initial grasps each scene. The numbers in Table~\ref{tab:all_planning_results} are averaged over 10 runs for each scene.
 
% The CHOMP planner \cite{ratliff2009chomp} is denoted as ``CHOMP'' and its goal set variant \cite{dragan2011learning} is denoted as ``CHOMP-C''. We also compare our algorithm with two sampling planners implemented in OMPL \cite{sucan2012open} embedded in Moveit! \cite{chitta2012moveit}: ``FMT'' \cite{janson2015fast} and ``RRT-Connect'' \cite{kuffner2000rrt}. 
 
Comparing OMG with other planners, we can see the improvement on planning and execution thanks to our grasp selection and refinement process. By modeling a probability distribution on the grasp set, our method globally selects the optimal grasp for planning. In our metrics, planning success acts as a feasibility and optimality of the trajectory but execution involves control and system noises. By terminating at the first feasible trajectory, other methods may generate sub-optimal plans to execute. In contrast, with the option to select goals in the goal set, our trajectory optimization has fundamentally more solutions and more likely to yield robust results. Some qualitative planning results are shown in Fig.~\ref{fig:planning_scenes}.

% While these methods admit a brute force search for all goals to find a global minimum, it wastes time in many scenarios.  Moreover, our planner OMG further refines the grasp quality of the goals to improve success compared with MD in Table. \ref{table1}.

\begin{figure}
    \centering
\begin{subfigure}{.25\textwidth}
  \centering
  \includegraphics[width=0.95\linewidth]{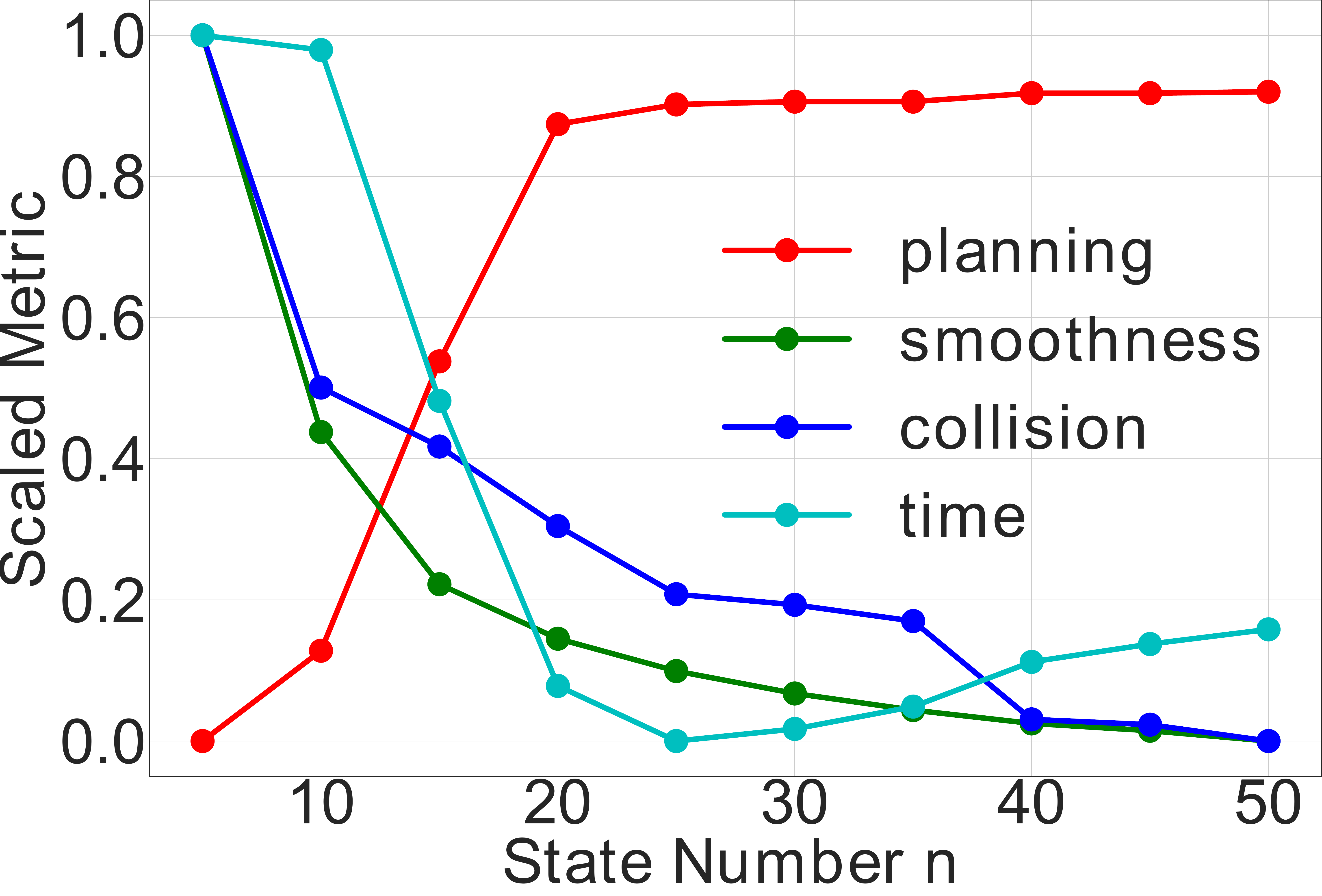}

  \label{fig:ablation1}
\end{subfigure}%
\begin{subfigure}{.25\textwidth}
  \centering
  \includegraphics[width=0.95\linewidth]{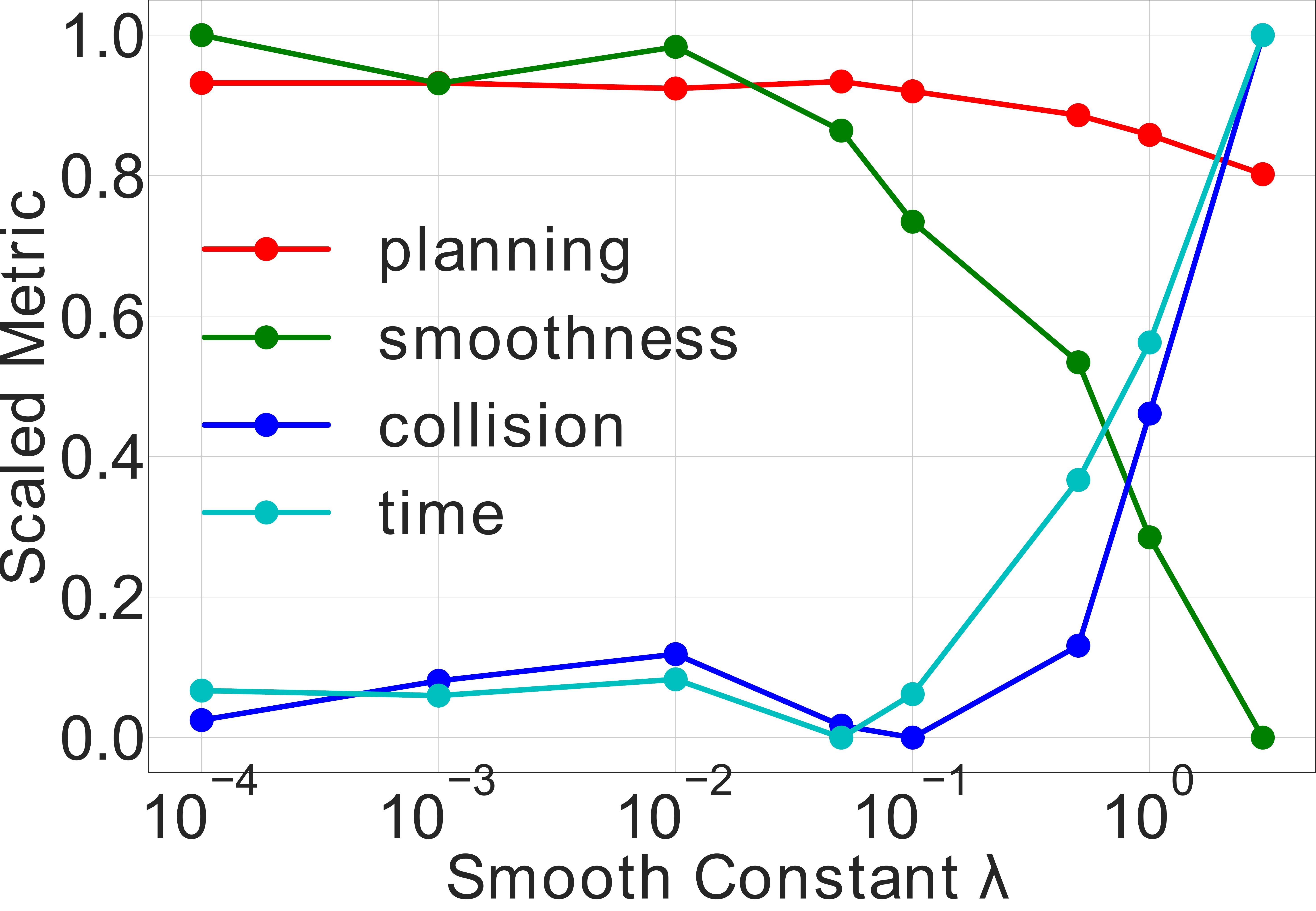}
  \label{fig:ablation2}
\end{subfigure}

\caption{Ablation study of $n$ and $\lambda$. The metrics (except planning) are scaled between $0$ and $1$.}
\label{fig:ablation}
\vspace{-4mm}
\end{figure}
	
% Left) We see decrease of smoothness and increase of success rate due to finer representation. The planning time drops first due to more solvable problems and then increases due to increase of computation. Right) The smoothness weight $\lambda$ around $0.1$ acts as a good balance in terms of collision and planning success.	

% \subsection{Ablation Study}  \label{sec:4d}

We also conduct ablation study on the influences of two important design choices in our OMG planner: the discretization $n$ defined in Sec.~\ref{sec:update_rule} and the smoothness constant $\lambda$ defined in Eq.~\eqref{eq:motion}. The results are presented in Fig.~\ref{fig:ablation}. By increasing $n$, a fine-grained trajectory better approximates a continuous curve and therefore induces better smoothness and planning success. The smoothness weight $\lambda$ around $0.1$ acts as a good balance in terms of collision and planning success.
% It is often desirable to gradually increase $\lambda$ during optimization \cite{ratliff2009chomp}, which is similar to dual gradient descent and intuitively enables a trajectory to curve around obstacles and then smooth itself.

% Since the functional gradient is based on calculus of variations that treats trajectory as continuous function. 

% Smoothness (defined in Sec. \ref{sec:chomp_obj}) is inherited from CHOMP and more heavily exploited in our planner. Specifically, we investigate the role of . 

\section{Conclusion}
\label{sec:conclusion}

We have presented a joint motion and grasp planning approach based on first-order trajectory optimization to generate grasping trajectories. Our OMG planner does not require a perfect goal configuration, since the candidate goals are determined during the planning process. A goal set is explicitly modeled as a probability distribution and the best goal is selected online. Moreover, our method synthesizes new grasps online to augment the goal set. Our experiments demonstrate that the OMG planner can plan grasping trajectories for cluttered scenes efficiently and robustly. While our work is based on CHOMP, we believe that the online goal selection and refinement procedure can also benefit other planning methods and tasks that involve a set of goals.

% Moreover, we make few assumptions about the structure of the grasping tasks, which implies that our framework could easily be extended to tasks that has a set of goal region.

% Overall, the algorithm integrates the solutions of the tasks traditionally needed for grasping an object: finding a feasible grasp, solving the inverse kinematic problem, and computing a collision-free trajectory. 

% Similar to other optimization-based planning methods, our approach is by nature local, and it has no guarantee to find feasible solutions. However, compared to sequential approaches, where a grasp is chosen from a set of pre-computed grasps, our method considers the whole set of feasible goal configurations in an online scheme, which is more efficient and principled. While our work is based on CHOMP, we believe the online goal selection procedure can also benefit other planning methods. Moreover, we make few assumptions about the structure of the grasping tasks, which implies that our framework could easily be extended to tasks that has a set of goal region. Our formulation also has close relationships with dynamic replanning and adaptive control. An interesting next step is to use the planning with full observation states to supervise method with perceptions.

%  We also plan to compare our methods with other sampling-based integrated planners \cite{hang2016hierarchical} \cite{vahrenkamp2010integrated} \cite{fontanals2014integrated} as well as compare the goal selection output with regression in \cite{dragan2011learning}.

%% Use plainnat to work nicely with natbib. 

\bibliographystyle{plainnat}
\bibliography{paper.bib}

\end{document}